\documentclass[]{spie}  
\usepackage[]{graphicx}
\usepackage{multirow}
\usepackage{caption}
\usepackage{subcaption}
\usepackage{amsmath}
\usepackage{hyperref}
\usepackage{amsfonts}
\usepackage{verbatim}
\usepackage{booktabs}
\usepackage{amsfonts}       
\usepackage{nicefrac}       
\usepackage{microtype}      
\usepackage{amsmath}
\usepackage[square,sort,comma,numbers]{natbib}
\usepackage{color}
\usepackage{algorithm}
\usepackage{algorithmic}
\DeclareMathOperator*{\argmin}{\arg\!\min}
\newcommand{\addpic}{8em}
\newcommand{\addpich}{10em}
\renewcommand{\tablename}{Figure}
\title{PADDIT: Probabilistic Augmentation of Data using Diffeomorphic Image Transformation} 

\author{Mauricio Orbes-Arteaga\supit{a,b,c,d,e}, Lauge S$\o$rensen\supit{a,b,c}, Jorge Cardoso\supit{d}, Marc Modat\supit{d}, Sebastien Ourselin\supit{d}, Stefan Sommer\supit{a}, Mads Nielsen\supit{a,b,c}, Christian Igel\supit{a}, Akshay Pai\supit{a,b,c}
\skiplinehalf
\supit{a}DIKU, University of Copenhagen, Denmark; \\
\supit{b}Biomediq A/S, Copenhagen, Denmark\\
\supit{c}Cerebriu A/S, Copenhagen, Denmark\\
\supit{d}King's College London, United Kingdom\\ 
\supit{e}University College London, United Kingdom
}

\authorinfo{\textit{Email addresses}: hmorbesa@biomediq.com (Mauricio), ap@cerebriu.com (Akshay Pai), madsn@di.ku.dk (Mads Nielsen)}

\begin{document} 
  \maketitle 

\begin{abstract}

For proper generalization performance of convolutional neural networks (CNNs) in medical image segmentation, the learnt features should be invariant under particular non-linear shape variations of the input. To induce invariance in CNNs to such transformations, we propose Probabilistic Augmentation of Data using Diffeomorphic Image Transformation (PADDIT) -- a systematic framework for generating realistic transformations that can be used to augment data for training CNNs. The main advantage of PADDIT is the ability to produce transformations that capture the morphological variability in the training data. To this end, PADDIT constructs a mean template which represents the main shape tendency of the training data. A Hamiltonian Monte Carlo(HMC) scheme is used to sample transformations which warp the training images to the generated mean template. Augmented images are created by warping the training images using the sampled transformations.  We show that CNNs trained with PADDIT outperforms CNNs trained without augmentation and with generic augmentation (0.2  and  0.15 higher dice accuracy respectively) in segmenting white matter hyperintensities from T1 and FLAIR brain MRI scans.

\end{abstract}



\section{Introduction}

Solving challenging image analysis tasks with machine learning -- especially, with convolutional neural networks (CNNs) -- requires a large amount of training data. However, in application areas such as medical image segmentation, the number of training patterns is typically very limited. In such cases, convolutional neural networks (CNNs) tend to overfit due to a lack of feature generalization to variations in shapes and appearance, and over parameterization. In order to address generalization, one has to find models that generate features equivariant or invariant under different transformations of the input. Equivariance of feature maps generated by CNNs to certain transformations can be obtained by using group convolutions \cite{cohen16} where different orientations of the features maps are learnt by kernels with shared weights. While group convolutions are very efficient due to weight sharing in learning multiple orientations for same feature maps, they are restricted to a limited set of transformations, i.e., symmetric, linear transformations. In order to reach generalization across a large group of transformations one has to rely on data augmentation. 


Data augmentation is commonly achieved by applying transformations that generate warped versions of the available training data. Accessing a larger group of transformations for augmentation is specially important in the field of medical image analysis because features related to the human anatomy need to maintain their identity under non-linear transformations. For instance, cortical surfaces of brain, structures with arbitrary shapes such as tumors, or structures subject to atrophy such as hippocampus in the brain have large variations in their expected morphology. The choice of the transformation in literature so far has been fairly arbitrary -- often restricted to rotations, translations, reflections, and very small nonlinear deformations~\cite{roth2015,NIPS2015_5854,hauberg2016}. Some degree of learning the right kind of transformations needed to improve the network performance was introduced in~\cite{NIPS2015_5854}. 
Hauberg et al.~\cite{hauberg2016} propose to learn a particular group of transformations. The authors suggest to use the space of transformations called diffeomorphisms, which are well-behaved in the sense of being differentiable and invertible, for transforming training data. In order to learn the kind of diffeomorphisms needed to account for all shape variations in the training data, the authors propose to measure relative shape changes by using non-linear image registration. From the resulting set of transformations, a distribution is constructed from which new transformations for augmentation are sampled using a Metropolis Markov chain Monte Carlo scheme (MMCMC). While the performance on MNIST \cite{lecun1998gradient} improved significantly, digits are simpler shapes compared to the more complex brain images considered in this study. Given the size of each brain  image, it is computationally intensive to randomly register sufficient pairs of images. In addition, since the posterior distribution of transformations is not a trivial space, MMCMC tend to get stuck in local isolated modes of distribution. Therefore, images that cannot be plausibly registered may induce transformations that are not meaningful.

In order to obtain a model that produces transformations that capture shape variations in training data automatically, we propose Probabilistic Augmentation of Data using Diffeomorphic Image Transformation (PADDIT). 
PADDIT involves an unsupervised approach to learn shape variations that naturally appear in the training dataset. This is done by first constructing an unbiased template image that represents the central tendency of shapes in the training dataset. We sample -- using a Hamiltonian Monte Carlo (HMC) scheme~\cite{duane1987} -- transformations that warp the training images to the generated mean template. The sampled transformations are used to perturb training data which is then used for augmentation. We use convolutional neural networks (CNNs) to segment T1/FLAIR brain magnetic resonance images (MRI) for white matter hyperintensities. We show that PADDIT outperforms CNN methods that use either no data augmentation or limited augmentation (using random B-spline transformations).

\section{Methods}
Probabilistic Bayesian models for template estimation in registration was introduced by \cite{zhang2013}, albeit using a different class of transformations. In short, the method views image registration as a maximum a posteriori (MAP) problem where the similarity between two images ($I_1$, $I_2$) is the likelihood. The transformations are (lie group exponential of a time-constant velocity field $\bold v$) regularized by a prior which is in the form of a norm attached to velocity field. Formally, it is a minimization of the energy
\begin{equation}
\label{reg}
E(I_1, I_2, \bold v) = \|I_1 \circ \mathrm{Exp}(\bold{v})  - I_2\|^2 + \lambda \|\bold{v}\|^2.
\end{equation}
The norm on the vector field is generally induced by a differential operator. However, we directly choose a  kernel {inducing a} reproducing kernel Hilbert space to parameterize the velocity field~\cite{Pai:2016dz}. Given a finite set of kernels, the regularization takes the form, 
\[
 \|\bold{v}\|^2 = \sum_i \sum_j a^T K(x_i,x_j) a,
\] 
where $a$ are the vectors attached to each spatial kernel, and $(x_{i},x_{j}) \in \Omega$ is the spatial position of each kernel $K$.

Using the $L2$ distance metric between two images (minimization of~\eqref{reg}), one can formulate template estimation as a Fr{\'e}chet mean estimation problem. In other words, given a set of $N$ images (or observations) $I_1, \dots, I_N$, the atlas $\hat{I} $ is the minimization of the sum-of-squared distances function 
\begin{equation}
\hat{I} = 	\argmin_{I_T} \frac{1}{N} \sum_{k=1}^N \|I_T-I_k\|^2.
\end{equation}

Since~\eqref{reg} is viewed as a MAP problem, the velocity fields are considered as latent variables, i.e., $a \sim \mathcal{N}(0, K)$, a normal distribution with zero mean and covariance $K$ derived from a kernel function. In the presence of latent variables, the template estimation is posed as an expectation maximization (EM) problem. Further, for simplicity, we assume an i.i.d.\ noise at each voxel, with a likelihood term (for each $k^{\text{th}}$ observation) given by
 \begin{equation}
p(I_k|\bold{v}_k,I_T,\sigma) = \frac{1}{(2\pi)^{V/2}\sigma^V} \mathrm{exp}\left(-\frac{\|I_T - I_k\circ\mathrm{Exp}(\bold{v}_k)\|^2}{2\sigma^2}\right),
\end{equation}
where  $\theta=\{ \sigma, I_T \}$ are the parameters to be estimated via MAP; $\sigma$ is the noise variance, $I_T$ is the mean template, and $V$ is the number of voxels. 
Each observation can be viewed as a random variation around a mean ($I_T \circ \mathrm{Exp}(- \bold v)$). The prior on the velocity field may be defined in terms of the norm as
\begin{equation}
p(\bold{v}_k) = \frac{1}{(2\pi)^{V/2}|K|^\frac{1}{2}}\mathrm{exp}\left(-\frac{\|\bold{v}_k\|^2}{2}\right)
\end{equation}

Estimating the posterior distribution involves the marginalization of it over the latent variables as
\begin{equation}
p(\theta|I_k) = p(I_k|\theta)p(\theta) = \int_\bold v p(I_k|\bold v,\theta) p(\bold v) d\bold v
\end{equation}

This is computationally intractable due to the dimensionality of $\bold v$. To solve this, Hamiltonian Monte Carlo (HMC)~\cite{neal2011} is employed to sample velocity field for marginalization. The posterior distribution to draw $S$ number of samples from is

\begin{multline}
\mathrm{log}\prod_{k=1}^N p(\theta|I_k) = \sum_{s=1}^S \mathrm{log}  \prod_{k=1}^N p(I_k|\bold{v}_{ks}, \theta) p(\bold{v}_{ks}), \\
\label{post1}
=\sum_{s=1}^{S} \left(-\frac{N}{2} \mathrm{log}~ |K| - \frac{1}{2}\sum_{k=1}^{N} a^T K a - \frac{MN}{2} \mathrm{log}~\sigma - \frac{1}{2\sigma^2}\sum_{k=1}^{N} \|I_k \circ \mathrm{Exp}(\bold{v}_k) - I_T\|^2~\right). 
\end{multline}

The sampled velocity fields ($\bold{v}_{ks}$ of the $k^{\text{th}}$ image) are used in an EM algorithm to estimate an optimal~$\theta$. The two steps are as follows:
\begin{itemize}
\item{\textbf{E-Step}: We draw samples from the posterior distribution~\eqref{post1} using HMC with the current estimate~$\theta_t$. Given $S$ sampled velocity fields, the mean is taken from:}
\begin{align}
\label{estep}
Q(\theta|\theta^t) =E_{\bold{v}_k|I_k,\theta_t} \left[- \sum_{k=1}^N \mathrm{log}~p(\theta|I_k)\right]
\end{align}
\item{\textbf{M-Step}: Update the parameters by maximizing $Q(\theta|\theta_t)$. The maximum form for $I_T$ and $\theta_t$ is given by:}
\begin{align}
\label{mstep}
\sigma^2 &= \frac{1}{MNS} \sum_{s=1}^{S}\sum_{k=1}^{N} \|I_T - I_k\circ \mathrm{Exp}(\bold{v}_{ks})\|^2\\
I_T &= \frac{ \sum_{s=1}^{S}\sum_{k=1}^{N} I_k \circ \mathrm{Exp}(\bold{v}_{ks}) |D\mathrm{Exp}(\bold{v}_{ks})|}{\sum_{s=1}^{S}\sum_{k=1}^{N}|D\mathrm{Exp}(\bold{v}_{ks})|}
\end{align}
\end{itemize}

A single-scale Wendland kernel~\citep{Pai:2016dz} is used to parameterize the velocity field and construct the covariance matrix for regularization. Once a template is estimated, the posterior distribution is sampled for a set of velocity fields for each training data. To induce more variations, the velocity fields are randomly integrated between 0 and 1. The training samples are deformed with cubic interpolation for the image, and nearest neighbor interpolation for the atlas to create the new set of synthetic data. The input (for one image as an example) to the deep-learning network will be of the form
\begin{equation}
\langle\langle I_n,L_n\rangle{},\langle I_n\circ\mathrm{Exp}(\bold{v}_{n1}), L_n\circ\mathrm{Exp}(\bold{v}_{n1}) \rangle{}, \dots \langle I_n\circ\mathrm{Exp}(\bold{v}_{nA}), L_n\circ\mathrm{Exp}(\bold{v}_{nA}) \rangle{} \rangle{},
\label{paddit}
\end{equation}
Where $A$ is the number of augmentations and $L_n$ is the label of input image $I_n$.
{Note that the label is a segmentation assigning a class to each voxel and is transformed using the same transformation accordingly.} 
Algorithm~\ref{Padditalgo} summarizes the workflow of PADDIT in pseudo-code.
\begin{algorithm}[H]
\caption{PADDIT}
  \begin{algorithmic}[1]
   \STATE Generate template using Equations~\eqref{estep} and \eqref{mstep}.
    \FOR{number of training epochs}
       \STATE Sample $A=2$ velocity fields per training image using HMC\cite{neal2011} from  the distribution~\eqref{post1}. 
       \STATE Integrate the sampled velocity field upto a randomly chosen time $t\le 1$ to warp the training image and its corresponding label image. 
       \STATE Extract slices from the warped images and add  them to the slices extracted from the original images, see~\eqref{paddit}.
       \STATE Train the convolutional neural network to classify each voxel.
 \ENDFOR
  \end{algorithmic}
\label{Padditalgo}
\end{algorithm}
\newcommand{\cp}{\ensuremath{\text{Cp}}}
\newcommand{\sd}{\ensuremath{\text{Sd}}}
\section{Experiments and Results}
We considered CNNs based on  a U-net architecture in our experiments. To evaluate the proposed method, the performance of  CNNs trained with data augmentation using PADDIT was compared to training without augmentation and training with augmentation using deformations based on random B-splines  -- we call this method the baseline. The above-mentioned strategies were applied to White Matter Hyperintensities (WMH) segmentation from FLAIR and T1 MRI scans. To this end, we use the training dataset from the 2017 WMH segmentation MICCAI challenge \footnote{http://wmh.isi.uu.nl}.  The set is composed of T1/FLAIR MRI scans and manual annotations for WMH from 60 subjects. Manual notations were performed in FLAIR space, therefore T1 modalities have been registered to such space. The images were also corrected for bias field inhomogeneities using SPM12. As further preprocessing images were cropped or padded to 200 $\times$ 200 $\times$ 200 voxels. Also, images were subtracted by its mean and divided by its variance, to normalized voxel intensities. The dataset was split into a training(30), validation(5) and testing(10) set. For each method two different deformed versions of each training case were created, i.e the training set size was tripled.  





The Random deformations for the baseline were obtained by using a deformation field defined on a grid with $\cp{}$ number of control points and B-spline interpolation. The size of deformation was controlled by adding Gaussian noise with $0$ mean and standard deviation $\sd{}$. We evaluate the impact of  $\cp{}$ and $\sd{}$ hyperparameters, specifically we tried: $\cp{} = [4\times4\times4,8\times8\times8,16\times16\times16]$ and $\sd{} = [2,4,6] $. 


Figure \ref{fig:ex_deformations} shows examples of the obtained deformed versions of a FLAIR scan from one subject from the training dataset.  As can be observed, both methods generated new shapes for WMHs regions.  It is worth noting, however,  that images provided by PADDIT look more realistic and without drastic alterations to the Brain. In contrast, those obtained using random B-spline deformations exhibit some aberrations in cortical and ventricular structures depending on the size of the deformation used.
\begin{table}[!ht]
\centering
\begin{tabular}{c c|cccc}
 & & &  \sd{}:2 & \sd{}:4 &  \sd{}:6 \\
\rotatebox{90}{ \ \ \ \ \ \ \  FLAIR} &\includegraphics[width=\addpic,height=\addpich,angle=90]{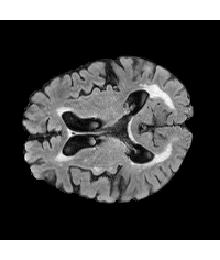} & 
\rotatebox{90}{\ \ \ \ \cp{}: 4} & 
\includegraphics[width=\addpic,height=\addpich,angle=90]{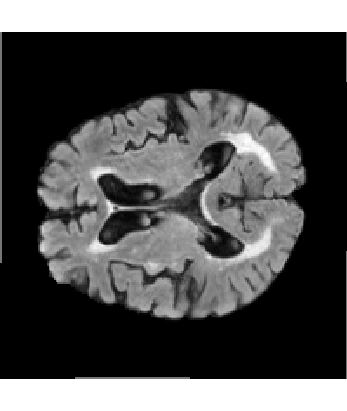} &
\includegraphics[width=\addpic,height=\addpich,angle=90]{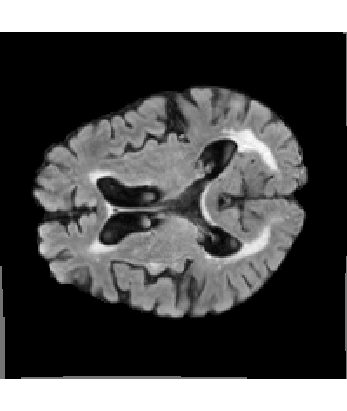} &
\includegraphics[width=\addpic,height=\addpich,angle=90]{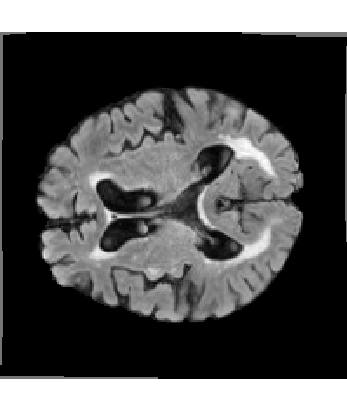}\\

\rotatebox{90}{\small{ \ \ \ \ PADDIT 1}} & \includegraphics[width=\addpic,height=\addpich,angle=90]{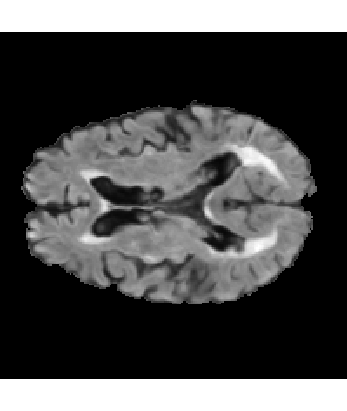} & 
\rotatebox{90}{\ \ \ \ \ \cp{}: 8} & 
\includegraphics[width=\addpic,height=\addpich,angle=90]{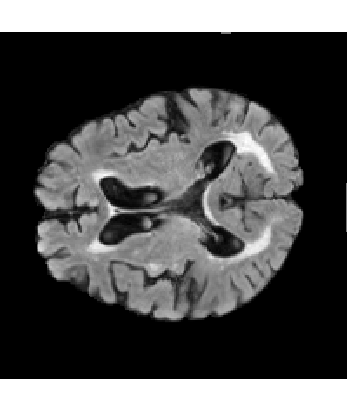} &
\includegraphics[width=\addpic,height=\addpich,angle=90]{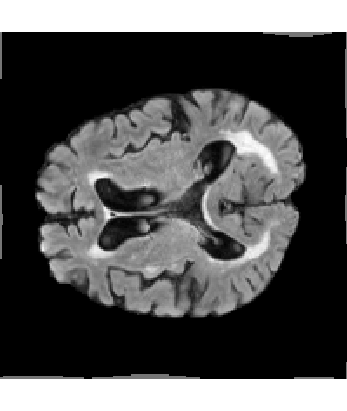} &
\includegraphics[width=\addpic,height=\addpich,angle=90]{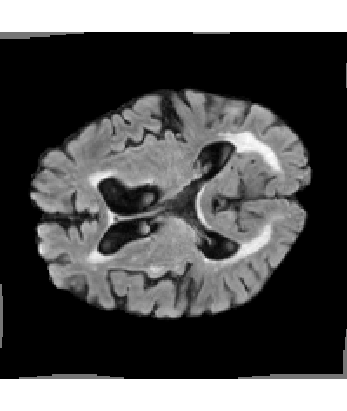}\\

\rotatebox{90}{ \ \ \ \ PADDIT 2} &\includegraphics[width=\addpic,height=\addpich,angle=90]{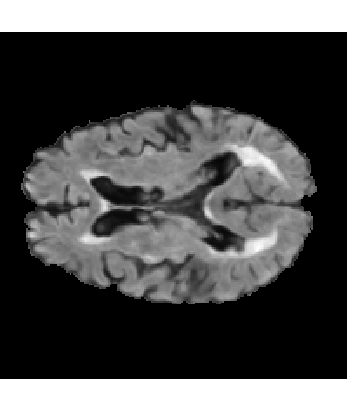} & 
\rotatebox{90}{\ \ \ \ \ \cp{}: 16} &
\includegraphics[width=\addpic,height=\addpich,angle=90]{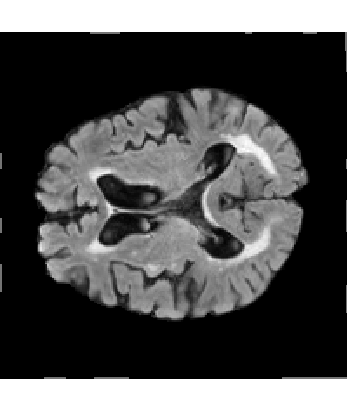} &
\includegraphics[width=\addpic,height=\addpich,angle=90]{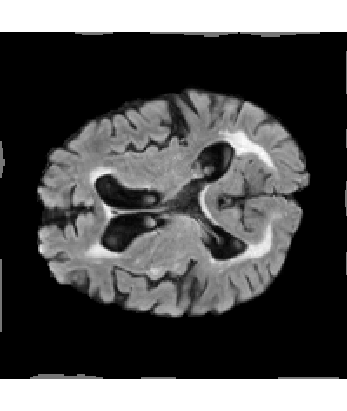} &
\includegraphics[width=\addpic,height=\addpich,angle=90]{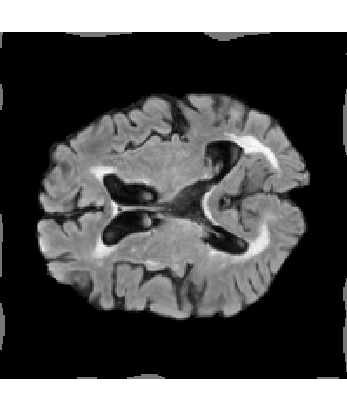}\\
\end{tabular}
\caption{Example of generated deformations. The first column shows the original FLAIR image, and the two deformed versions using PADDIT. Remaining columns show different configurations used to get the random B-spline based deformations}\label{fig:ex_deformations}
\end{table}

As we mentioned before, we split the data into validation and training set. In this case the validation set is used to tune the best configuration for the number of control points $\cp{}$ and size of deformation $\sd{}$ . Figure \ref{fig:Performance_valtest}, shows the dice performance at each epoch on the validation and testing set. It is worth noting that PADDIT achieved higher accuracy than training with random B-spline deformations as well as  training without augmentation. Also, it can be noted that random B-spline deformations did not provide a consistent improvement compared to the training without data augmentation.

\begin{table}[!ht]
\centering
\begin{tabular}{cc}
\includegraphics[height=7cm,width=8.5cm]{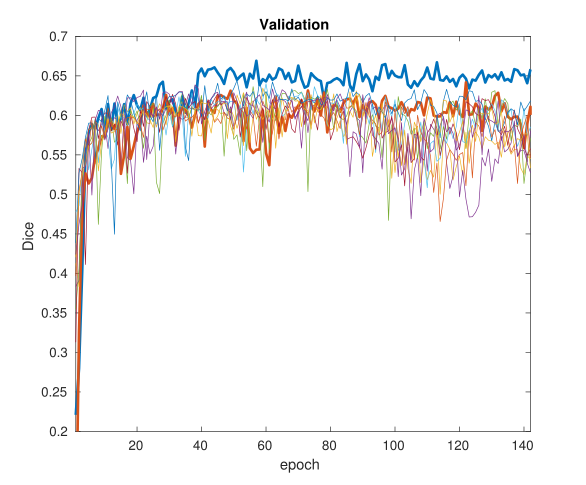} & \includegraphics[height=7cm,width=8.5cm]{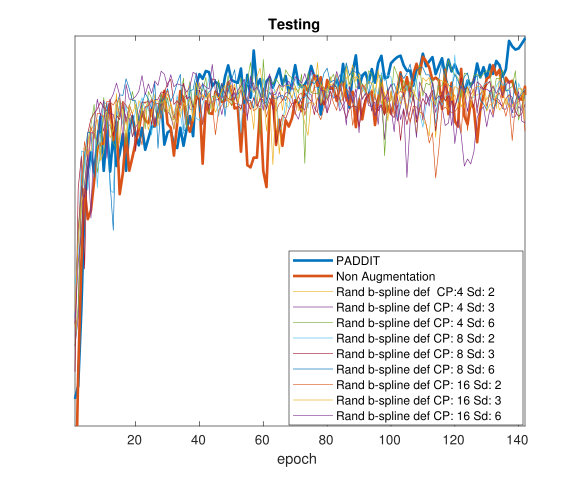} \\
\end{tabular}
\caption{Performance on the validation and testing set for each method. Dice is computed at each epoch}\label{fig:Performance_valtest}
\end{table} 

For the final assessment of PADDIT, the validation data was used for early stopping. The final evaluation of each method is carried out on the testing set using the network configuration at the epoch where it showed the highest accuracy on the validation set. The best configuration for random deformations was achieved using $\cp{} = 8\times 8 \times 8$ and  $ \sd{} = 4$  For PADDIT, the control points were placed every 8 voxels. Results for evaluation on the testing set are summarized in Table \ref{Final_evaluation}. Our proposed method PADDIT achieved $\approx 0.2$ higher dice accuracy compared to the network performance without data augmentation and $\approx 0.15$ compared to the baseline data augmentation approach (best configuration). (both differences where statistically significant ($p<0.5$))

\renewcommand{\tablename}{Table}
\begin{table}[!ht]
\centering
\begin{tabular}{cccccccccccc}
\hline
 & \rotatebox{90}{\vtop{\hbox{\strut Non Data} \hbox{\strut Aug}}} &  \rotatebox{90}{\vtop{\hbox{\strut Rd \cp{}: 4} \hbox{\strut \sd{}:2}}} &  \rotatebox{90}{\vtop{\hbox{\strut Rd \cp{}: 4} \hbox{\strut \sd{}:4}}} &  \rotatebox{90}{\vtop{\hbox{\strut Rd \cp{}: 4} \hbox{\strut \sd{}:6}}} &  \rotatebox{90}{\vtop{\hbox{\strut Rd \cp{}: 8} \hbox{\strut \sd{}:2}}} &  \rotatebox{90}{\vtop{\hbox{\strut Rd \cp{}: 8} \hbox{\strut \sd{}:4}}} &  \rotatebox{90}{\vtop{\hbox{\strut Rd \cp{}: 8} \hbox{\strut \sd{}:6}}} &  \rotatebox{90}{\vtop{\hbox{\strut Rd \cp{}: 16} \hbox{\strut \sd{}:2}}} &  \rotatebox{90}{\vtop{\hbox{\strut Rd \cp{}: 16} \hbox{\strut \sd{}:4}}} &  \rotatebox{90}{\vtop{\hbox{\strut Rd \cp{}: 16} \hbox{\strut \sd{}:6}}} & \rotatebox{90}{PADDIT } \\
\hline 
\hline
Dice (mean) $(\%)$ &   66.32   &  66.28  & 63.47 & \underline{66.61}& 64.52 & 64.38 & 65.66  & 63.58 & 65.87 & 65.35 & \textbf{68.13} \\
Dice (std) $(\%)$ &   24.82   &  22.60  & 24.66 & \underline{22.74} & 23.47 & 24.03 & 23.27  & 24.57 & 22.38 & 23.41 & \textbf{21.85} \\
\hline
\end{tabular}
\caption{Segmentation accuracy for all the assessed strategies, the highest dice score achieved by the random B-spline deformation approach is underlined}\label{Final_evaluation}
\end{table}

\section{New or breakthrough work to be presented}
Even though several configurations of random transformations generated realistic looking images, they were not necessarily useful in CNN training. On the other hand, the best configuration of random transformations generated images that were not necessarily biologically plausible. We hypothesize that such noisy data may help the optimization to find better minimums. However, one has to be careful in choosing the configuration of transformations since other configurations with a higher magnitude of deformations had a negative effect on the training. In the case of PADDIT, one need not worry about the transformation configuration too much since the method learns the right transformation needed to capture the shape variations in the data set. Hence, the resulting synthetic images were both realistic and useful for CNN training.
\section{Conclusion}

In this paper, a probabilistic data augmentation approach using diffeomorphic image transformations is proposed. Contrary to traditional augmentation strategies that used predefined or aleatory transformations the proposed method is able to learn transformations that better capture the anatomical variations of the training dataset, while the structural topology is preserved. The proposed probabilistic augmentation approach PADDIT, proved to be an effective way to increase the training set by generating new training images which improve the segmentation performance of CNN's based approaches. From the results, it is evident that the network trained with is performed statistically significantly better than the networks with either no data augmentation or random B-splines based augmentation. 


\section*{Acknowledgments}
This project has received funding from the European Union’s Horizon 2020 research and innovation
programme under the Marie Skłodowska-Curie grant agreement No 721820. We would like to thank both Microsoft and NVIDIA for providing computational resources on the Azure platform for this project.

\bibliographystyle{spiebib} 
\bibliography{report}   

\end{document}